# Nichtverbales Verhalten sozialer Roboter

**Bewegungen, deren Bedeutung und die Technik dahinter**

*Kathrin Janowski, Elisabeth André*

*Universität Augsburg, Germany*

„M**ache die Figuren in solchen Geberden, dass diese zur Genüge zeigen, was die Figur im Sinn ha**t. Wo nicht, so ist deine Kunst nicht lobenswerth." - Leonardo da Vinci[1]

**Zusammenfassung**

Nichtverbale Signale sind ein elementarer Bestandteil der menschlichen Kommunikation. Sie erfüllen eine Vielzahl von Funktionen bei der Klärung von Mehrdeutigkeiten, der subtilen Aushandlung von Rollen oder dem Ausdruck dessen, was im Inneren der Gesprächspartner vorgeht. Viele Studien mit sozial-interaktiven Robotern zeigen, dass vom Menschen inspirierte Bewegungsmuster ähnlich interpretiert werden wie die von realen Personen. Dieses Kapitel erläutert daher die wichtigsten Funktionen, welche die jeweiligen Bewegungsmuster in der Kommunikation erfüllen, und gibt einen Überblick darüber, wie sie auf Roboter übertragen werden können.

## 1. Einleitung

Ein großer Teil der menschlichen Kommunikation findet auf der nonverbalen Ebene statt. Ein Lächeln, ein Nicken oder eine Handbewegung können oft Botschaften übermitteln, für welche die Sprache einen ganzen Satz benötigt. Besonders im sozialen Umgang mit anderen Menschen spielt es nicht nur eine Rolle, welche Worte verwendet werden, sondern auch, was der Rest des Körpers tut. Etliche Kursangebote bereiten Redner und Arbeitssuchende darauf vor, wie sie sich vor Publikum oder potenziellen Arbeitgebern bewegen sollten, um den besten Eindruck zu hinterlassen. Personen, die einander gut kennen, verstehen sich ohne Worte, und Pokerspieler erkennen am Zucken der Gesichtsmuskeln, wie gut das Blatt des Mitspielers ist. Die meisten dieser Signale werden unbewusst gesendet und interpretiert, oft schon von Geburt an (Knapp et al. 2013).

Nicht erst seit der Pixar-Lampe aus dem Kurzfilm *Die kleine Lampe* (1986) wissen wir, wie leicht Menschen auch in die Bewegungen unbelebter Objekte Emotionen, Persönlichkeit oder Denkprozesse hineininterpretieren. Das macht die Körpersprache immens wichtig für die Kommunikation mit Maschinen, welche nicht nur technisch funktionieren, sondern darüber hinaus eine soziale Rolle einnehmen sollen. Gerade in Bereichen, in denen Alltagsnutzerinnen und -nutzer mit Robotern sozial interagieren, ist eine Orientierung an menschlichen Verhaltensweisen sinnvoll und kann dazu beitragen, den Zugang zu Technologie und deren Akzeptanz verbessern. Dies gilt u.a. für Roboter in der Rolle von Gedächtnistrainern, Gesundheitsberatern oder persönlichen Assistenten (Janowski et al. 2018). Dieses Kapitel befasst sich deshalb damit, welche Bewegungsmuster der Mensch über Jahrtausende hinweg einstudiert hat und wie diese auf sozial-interaktive Roboter übertragbar sind.

Der folgende Abschnitt wird zunächst einige allgemeine Konzepte erläutern, welche die Funktion der nichtverbalen Kommunikation betreffen. Danach folgen mehrere Abschnitte, welche sich verschiedenen Klassen von Bewegungen widmen, grob orientiert an den beteiligten Körperteilen. Dabei wird zuerst erläutert, welche Funktion die Bewegungen bei Menschen und Robotern erfüllen, und anschließend ein Überblick über die technische Umsetzung in der Robotik gegeben. Zuletzt wird noch ein Abschnitt auf die Koordination von Bewegungen und gesprochener Sprache eingehen, bevor das Kapitel mit einer abschließenden Zusammenfassung endet.

## 2. Allgemeine Grundlagen

Bei der Entwicklung von Verhaltensweisen für sozial-interaktive Roboter orientiert man sich üblicherweise an Untersuchungen zu menschlichen Verhaltensweisen aus den Sozialwissenschaften.

---

[1] „Das Buch von der Malerei", Übersetzung von Heinrich Ludwig, 1882; Wilhelm Braumüller, Wien

Häufig werden auch Daten von Menschen in sozialen Interaktionen statistisch ausgewertet und als Grundlage für die Umsetzung des nichtverbalen Verhaltens von sozial-interaktiven Robotern verwendet. Aus diesem Grund beschreiben wir im Folgenden zunächst relevante Vorarbeiten aus den Verhaltenswissenschaften und untersuchen im Anschluss, inwieweit diese durch Roboter nachgeahmt werden können.

## 2.1 Gemeinsamer Redehintergrund

Jede gemeinsame Handlung setzt voraus, dass sich die beteiligten Personen über den Inhalt und Ablauf der Handlung einig sind (Clark und Brennan 1991). Um ein Gespräch zu führen, müssen beide Teilnehmer nicht nur wissen, was das Thema ist, sondern auch sicherstellen, dass sie die Aussagen und Fragen des Gegenübers gehört und richtig interpretiert haben.

Viele der dafür notwendigen Bestätigungen finden auf der nichtverbalen Ebene statt. Kurze Signale wie Kopfnicken (Clark und Brennan 1991), Blickkontakt (Argyle und Graham 1976, Clark und Brennan 1991) und der Blick auf relevante Objekte in der Umgebung (Argyle und Graham 1976) zeigen (s. Abb. 1 und Abb. 2), ob der Zuhörer aufmerksam ist und den Sprecher verstanden hat. Sprache wird oft mit Handgesten kombiniert, um möglichst effizient eine unmissverständliche Aussage zu treffen (Clark und Krych 2003, Kendon 2004, van der Sluis und Krahmer 2007).

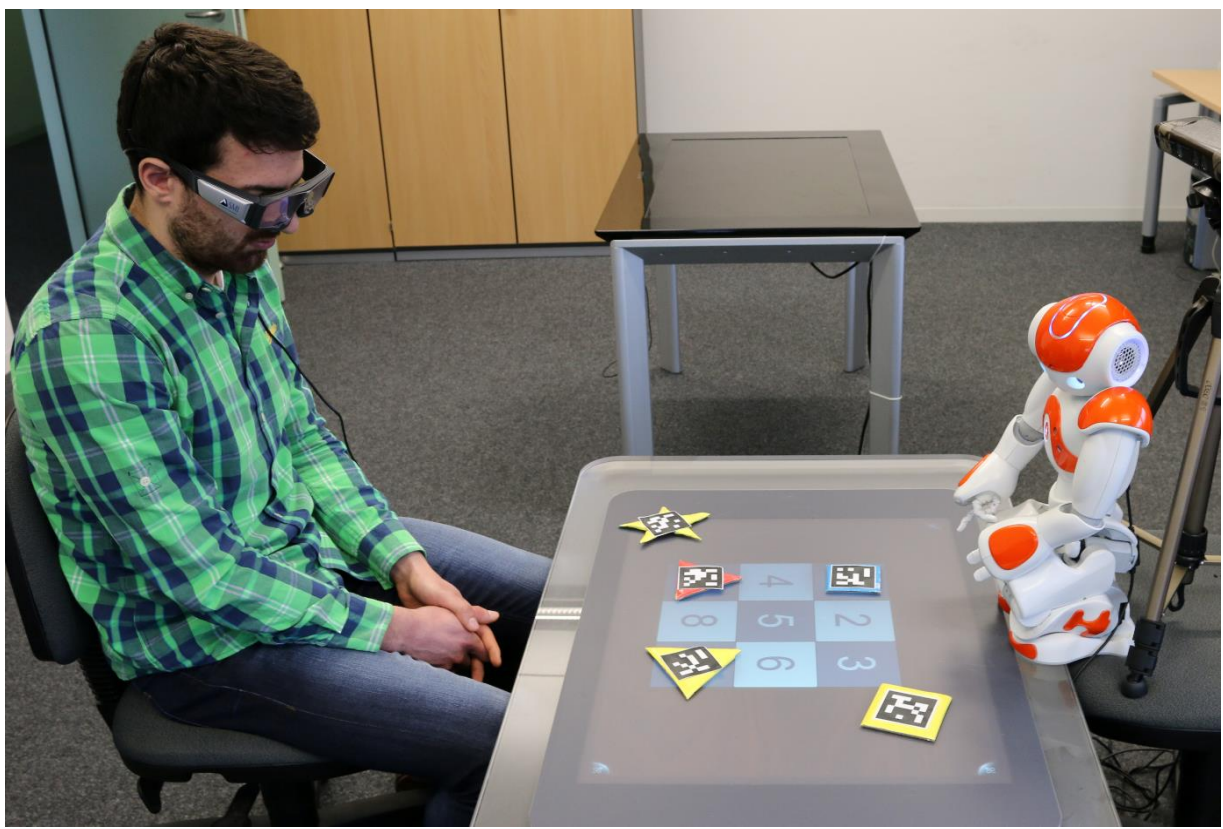

Abb. 1: Aufmerksamkeit von Nutzer und Roboter liegen auf dem gleichen Objekt.

## 2.2 Rederechtsvergabe

Eine spezielle Form des gemeinsamen Redehintergrunds ist die Aushandlung der Sprecher- und Zuhörerrollen. Dabei sollen einerseits Überlappungen vermieden werden, welche das Zuhören erschweren, und andererseits möglichst wenig störende Pausen entstehen. Auch hierbei wird die notwendige Koordination meistens auf die nichtverbalen Kanäle ausgelagert.

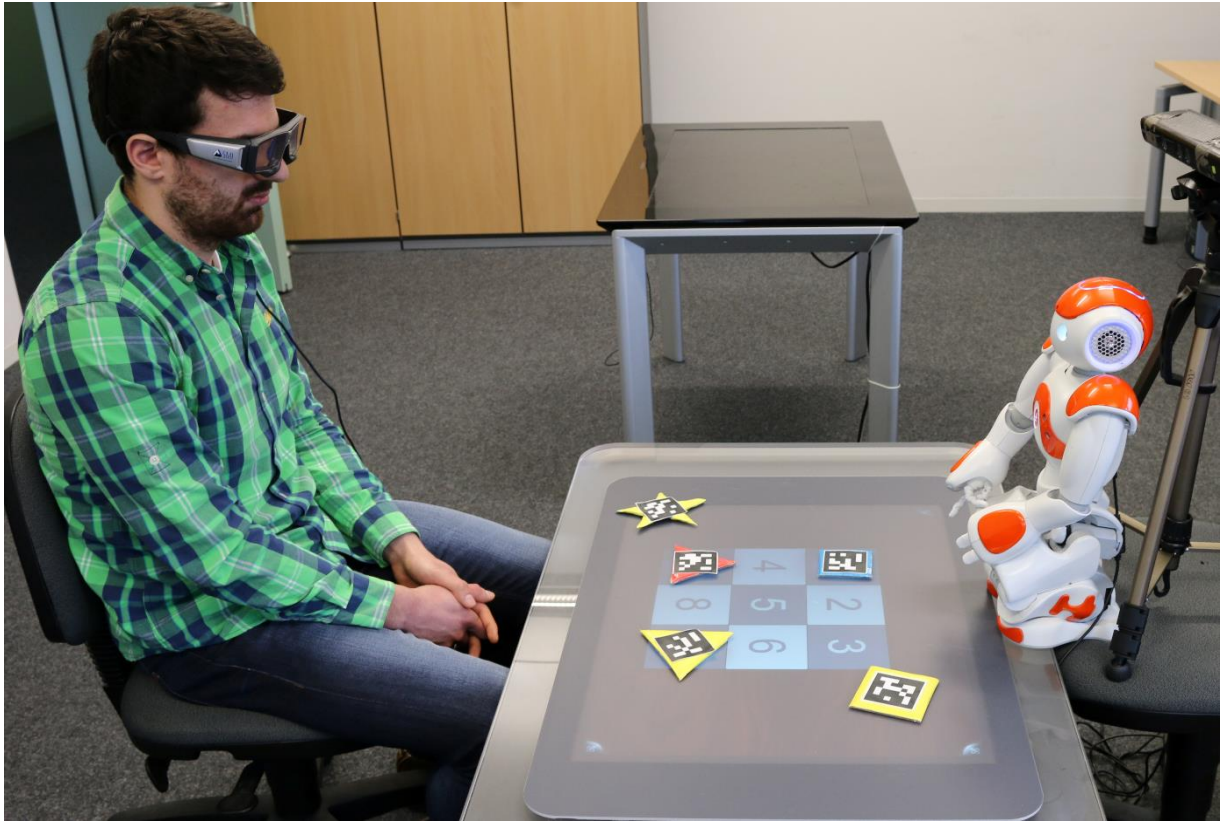

Abb. 2: Blickkontakt zwischen Nutzer und Roboter.

Vor allem das Blickverhalten zeigt, wie viel Aufmerksamkeit die Gesprächspartner einander widmen und ob sie aktuell bereit sind, Informationen zu empfangen (Argyle und Cook 1976). Aber auch andere Gesten können hier zum Einsatz kommen, beispielsweise das Heben einer Hand, um die Sprechabsicht anzuzeigen (Knapp et al. 2013).

## 2.3 Beziehung und Bindung

Körperhaltung, Blickrichtung oder Abstände können ausdrücken, wie jemand – im wahrsten Sinne des Wortes – zu anderen Personen steht. Auch hier spielt der Aspekt der Aufmerksamkeit eine wichtige Rolle. Freunden, Bekannten und Familienmitgliedern wird üblicherweise mehr Aufmerksamkeit zuteil als Fremden, denen man an der Bushaltestelle begegnet. Entsprechend selten nimmt man mit Fremden Blickkontakt auf (Argyle und Cook 1976).

Ein wichtiges Phänomen hierbei ist auch das Widerspiegeln von Mimik und Körperhaltung des Gesprächspartners, das auf Gruppenzugehörigkeit, Zuneigung oder Mitgefühl hindeuten kann (Knapp et al. 2013). Beispielsweise kommt es vor, dass Gesprächspartner die Beine ähnlich übereinanderschlagen oder Beobachter das Gesicht auf ähnliche Weise verziehen, wenn eine andere Person Schmerzen erleidet.

## 2.4 Persönlichkeit und Gefühlslage

Die aktuelle Gefühlslage eines Menschen wird selten verbal ausgedrückt. Die wenigsten Leute sagen explizit, wenn sie traurig oder wütend sind, sondern lassen ihren Tonfall, Gesicht und Körper sprechen. Ein anschauliches Beispiel dafür, wie stark nichtverbale Kommunikation hier bevorzugt wird, sind die in Chat-Anwendungen verbreiteten Emojis und Emoticons.

Bei „Emotionen" handelt es sich um kurzfristige Gefühlsregungen, die sich oft nur für wenige Sekunden im Gesicht zeigen (Ekman und Friesen 2003). Zur „Persönlichkeit" zählen dagegen allgemeine Verhaltensmuster, die sich nur langsam über das Leben hinweg verändern. Bei beidem gibt es weit verbreitete Annahmen, wie sich diese in der Körpersprache äußern (Knapp et al. 2013). Da

Gefühlsregungen und Persönlichkeit schwer objektiv zu beurteilen sind, sollte deren Zuordnung zur Körpersprache allerdings mit Vorsicht behandelt werden (Knapp et al. 2013; Feldman Barrett et al. 2019).

## 2.5 Das „unheimliche Tal"

Eine gängige Theorie zur Gestaltung von Robotern besagt, dass diese mit zunehmender Ähnlichkeit zum Menschen zunächst immer positiver wahrgenommen werden, aber die Situation drastisch kippt, wenn die Ähnlichkeit zu groß wird (Mori et al. 2005). Bei Robotern, welche beinahe menschlich wirken, fallen Fehler aufgrund der höheren Erwartung umso stärker ins Gewicht, sodass sie schnell gruselig und abstoßend wirken. Die steigende Zuneigung sinkt also bildlich gesprochen in ein tiefes Tal ab, kurz bevor sie den Gipfel erreicht. Entsprechend ist dieses Phänomen als „Uncanny Valley" bekannt.

Bewegungen haben in diesem Zusammenhang einen verstärkenden Effekt. Einerseits wirken Maschinen, welche lebensähnliche Bewegungen zeigen, schneller liebenswert als regungslose Apparate, aber andererseits ist die Ablehnung bei unpassenden Bewegungen umso tiefer (Mori et al. 2005). Um dieses Problem zu umgehen, empfiehlt es sich oft, entweder das Aussehen des Roboters oder dessen Bewegungen absichtlich mechanisch zu gestalten.

## 3. Konkrete Bewegungsmuster

Die folgenden Abschnitte betrachten, welche Rolle die jeweiligen Körperteile in der nichtverbalen Kommunikation spielen. Dabei wird zunächst erklärt, welche der zuvor genannten Funktionen mit den entsprechenden Bewegungen im Zusammenhang steht. Es wird beleuchtet, wie die Bewegungsmuster zwischen Menschen eingesetzt und interpretiert werden, und anhand von Beispielen gezeigt, wie sich diese Erkenntnisse auf sozial-interaktive Roboter übertragen lassen. Im Anschluss daran wird die technische Umsetzung der Bewegung für Roboter näher erläutert.

## 3.1 Blickrichtung

Menschen sind stark visuell orientiert. Sie nutzen nicht nur die eigenen Augen für einen Großteil der Wahrnehmung, sondern reagieren auch stark auf die Augen von anderen Personen (Argyle und Cook 1976). Entsprechend wichtig sind die Augen bei der Gestaltung von sozial-interaktiven Robotern. Dieser Abschnitt erklärt, welche Rolle das Blickverhalten in der Kommunikation einnimmt und wie es auf Robotern umgesetzt wird.

### *3.1.1 Funktionen der Blickrichtung*

Um Informationen zu erhalten, ist es oft notwendig, einen bestimmten Gegenstand zu betrachten, beispielsweise um dessen Farbe zu erkennen oder eine Beschriftung zu lesen. Auch Gesprächspartner werden genau beobachtet, um deren Körpersprache wahrnehmen und Schlüsse daraus ziehen zu können. Umgekehrt verrät die Richtung, in die der Partner blickt, wen oder was er in diesem Moment für wichtig hält. Im Fall eines sozial-interaktiven Roboters dienen Blicksignale dazu, die Absichten des Roboters für den Menschen nachvollziehbar zu machen.

Bereits Kleinkinder folgen dem Blick anderer Personen oder versuchen, diesen auf interessante Objekte zu lenken (Mundy und Newell 2007). Um zu erkennen, ob die andere Person verstanden hat, blicken Menschen häufig zurück zu ihrem Gegenüber und prüfen, ob deren Blick auf dem richtigen Objekt ruht (Clark und Krych 2003; Mundy und Newell 2007).

Diese Blickmuster verbessern auch die Mensch-Roboter-Interaktion. Wenn ein Roboter auf aktuell relevante Objekte blickt, ist für Menschen leichter nachvollziehbar, ob der Roboter ihre Anweisung korrekt verstanden hat, und der Roboter wird entsprechend als intelligenter wahrgenommen (Huang und Thomaz 2011). Dieses Verhalten wirkt außerdem natürlicher und vermittelt den Eindruck, dass der Roboter stärker an der Zusammenarbeit interessiert ist (Mehlmann et al. 2014). Beobachter bevorzugen außerdem Roboter, welche beim Übermitteln von Informationen wiederholt zum Gesprächspartner blicken, als ob sie dessen Aufmerksamkeit prüfen wollten (Huang und Thomaz 2011).

Die visuelle Aufmerksamkeit spielt auch eine wesentliche Rolle bei der Rollenverteilung im Gespräch. Wer zum Partner blickt, signalisiert damit die Bereitschaft, Nachrichten auf verbaler und nichtverbaler Ebene zu empfangen (Argyle und Cook 1976). Sprecher neigen deswegen dazu, den Blick vom Partner abzuwenden, während sie mit der Planung ihres eigenen Satzes beschäftigt sind. Umgekehrt suchen sie nach Teilsätzen Blickkontakt, um zu überprüfen, ob das bisher Gesagte verstanden wurde, oder sehen am Ende ihres Beitrags die Person an, welche als nächstes sprechen soll (Knapp et al. 2013).

In einer Studie mit dem Roboter NAO (Andrist et al. 2014) konnte gezeigt werden, dass ein Roboter, der zu Beginn seines Satzes den Blick abwendet, beim Menschen den Eindruck erweckt, als würde er genauer über seine Antwort nachdenken. Entsprechend später ergriffen Probanden das Wort, wenn der Roboter lange Zeit still blieb. Zu ähnlichen Ergebnissen kam auch eine Studie, in welcher der Roboter Furhat einem Menschen Zeichenanweisungen gab (Skantze et al. 2014).

Auch die Blicksignale, mit denen das Wort an andere Teilnehmer übergeben wird, lassen sich von Menschen auf sozial-interaktive Roboter übertragen. Mutlu et al. variierten in einer Studie die Blickmuster eines Roboters, um jeweils zwei Studienteilnehmer unterschiedlich stark in das Gespräch einzubeziehen (Mutlu et al. 2012). Teilnehmer, die er zum Ende des Satzes ansah, ergriffen fast immer das Wort, während diejenigen, welche kaum angesehen wurden, wenig zum Gespräch beitrugen.

*3.1.2 Umsetzung der Blickbewegung*

Für das Abwenden des Blicks ist es üblicherweise ausreichend, bestimmte Winkel für die Halsgelenke des Roboters vorzugeben, die von der Neutralstellung abweichen (Andrist et al. 2014). Um Blickkontakt zum menschlichen Gegenüber aufzunehmen, sind eine Kamera und Software zur Gesichtserkennung notwendig. Idealerweise ist diese Kamera in oder nahe an den Augen des Roboters platziert. Durch Kopfdrehungen lässt sich die Kamera schließlich so ausrichten, dass das Gesicht des Menschen im Zentrum des Kamerabilds liegt (Huang und Thomaz 2011). Für andere Objekte im Arbeitsraum ist es oft erforderlich, deren Koordinaten zu bestimmen und die entsprechenden Winkel über trigonometrische Verfahren zu berechnen.

Schwieriger sind die Bestimmung der geeigneten Zeitpunkte und der Wahrscheinlichkeiten, mit denen der Roboter auf ein bestimmtes Ziel blicken soll. Hierfür kommen häufig statistische Verfahren zum Einsatz, bei denen zunächst ermittelt wird, wie oft Menschen im Gespräch wohin blicken (Mutlu et al. 2012; Andrist et al. 2014). Wie lange ein Roboter dem Menschen ins Gesicht blicken darf, ist stark vom kulturellen Hintergrund abhängig und muss im Zweifelsfall gezielt in einer Studie ermittelt werden.

## 3.2 Gestik

Hand- und Kopfbewegungen treten sehr häufig in der zwischenmenschlichen Kommunikation auf, entweder als Ergänzung zur gesprochenen Sprache oder als vollwertiger Ersatz. Sie sind so stark im menschlichen Verhalten verankert, dass Menschen sogar nicken und gestikulieren, wenn der Gesprächspartner sie nicht sehen kann, beispielsweise am Telefon (Knapp et al. 2013). Besonders bei Robotern, deren Kopf oder Körper ohnehin menschenähnlich gestaltet ist, bietet es sich daher an, Botschaften auf diesem Weg zu übermitteln.

*3.2.1 Funktionen von Gesten*

Man kann grob zwischen sprachunabhängigen und sprachbezogenen Gesten unterscheiden (Knapp et al. 2013). Erstere beinhalten sogenannte *Embleme*, deren Bedeutung stark vom kulturellen Hintergrund der Beteiligten abhängt. Beispiele dafür sind etwa Kopfnicken, Winken zur Begrüßung oder der nach oben gestreckte Daumen als positive Bestätigung.

Zu den sprachbezogenen Gesten gehören beispielsweise solche, die konkrete oder abstrakte Konzepte veranschaulichen. McNeill unterscheidet hier zwischen *ikonischen* und *metaphorischen Gesten* (McNeill 1992). Erstere ahmen optische Merkmale von Gegenständen oder Bewegungen nach. Dadurch werden beispielsweise räumliche Informationen ergänzt oder verdeutlicht, die sich verbal nur sehr umständlich ausdrücken lassen. Letztere umfassen etwa das Formen symbolischer Behälter für Ideen oder das Anzeigen einer räumlichen Ausdehnung für ein weit gefasstes Thema (Knapp et al. 2013).

*Zeigegesten* deuten auf Objekte oder auch abstrakte Punkte im Raum, denen im Gespräch eine Bedeutung zugewiesen wurde (McNeill 1992). Diese Gesten können mit der ganzen Hand, dem ausgestreckten Zeigefinger oder auch durch ein Kopfnicken in die entsprechende Richtung ausgeführt werden. Durch Zeigegesten lassen sich sprachliche Verweise auf Objekte oft stark vereinfachen (Clark und Krych 2003).

Eine weitere Unterkategorie sind *Rhythmusgesten.* Diese dienen unter anderem dazu, einzelne Worte oder Satzteile zu betonen, oder den Inhalt zu strukturieren (Knapp et al. 2013).

In den meisten Fällen lassen sich diese Gesten von der zwischenmenschlichen Kommunikation auf die Mensch-Roboter-Interaktion übertragen. So existieren beispielsweise Studien zu Robotern, welche Zeigegesten einsetzen (Huang und Thomaz 2011; Sauppé und Mutlu 2014), oder Ansätze, um gleichartige Gestik sowohl auf menschenähnlichen virtuellen Charakteren als auch humanoiden Robotern darzustellen (Le und Pelachaud 2012).

*3.2.2 Umsetzung der Gestik*

Eine Geste besteht grundsätzlich aus drei Phasen: Vorbereitung, Kern (auf Englisch „stroke") und Rückzug (Kendon 2004). Der Kern ist dabei die Phase, welche die Bedeutung der Geste enthält. Vor und nach der Kernphase kann sich außerdem eine Haltephase befinden, während der die Hand in der entsprechenden Position verweilt.

Für die Umsetzung der Gestik bei Robotern empfiehlt es sich, nur die Kernphase festzulegen und die Übergangsphasen automatisch von der aktuellen Armposition abzuleiten (Le und Pelachaud 2012).

Dazu wird eine Reihe von Schlüsselpositionen definiert, welche nacheinander von den Motoren des Roboters angefahren werden. Besonders für Gesten wie Winken oder betonende Schläge, welche keinen Bezug zur Umgebung haben, werden diese Schlüsselpositionen meistens von einem Gestalter ermittelt und in einer Animationsdatei abgespeichert.

Andere Situationen erfordern dynamisch erzeugte Bewegungen, etwa für Zeigegesten zu einem Gegenstand, der vom Nutzer an beliebige Orte bewegt werden kann. In diesem Fall müssen die Winkel für die Gelenke abhängig von der gewünschten Handposition berechnet werden. Die so genannte „Inverse Kinematik" stützt sich dafür auf das Wissen über die Abstände zwischen den Gelenken, deren minimale und maximale Winkel und eventuelle Einschränkungen wie die Richtung, in die der Ellenbogen zeigen soll.

Sind die Schlüsselpositionen bekannt, müssen noch die Geschwindigkeiten für die Bewegung ermittelt werden. Dazu ist einerseits zu berücksichtigen, wie schnell jeder Motor maximal beschleunigt werden darf, um Schäden zu vermeiden. Um Verschleiß vorzubeugen, wird üblicherweise über mehrere hundert Millisekunden hinweg schonend beschleunigt und abgebremst. Langsamere Bewegungen haben außerdem den Vorteil, dass die Motorengeräusche weniger laut ausfallen.

Da Roboter – anders als virtuelle Charaktere – an physikalische Gesetze gebunden sind, sollten ausladende oder ruckartige Bewegungen vermieden werden, welche ihren Körper aus dem Gleichgewicht bringen können.

## 3.3 Körperhaltungen

Die Haltung des Körpers ist eng mit der Gestik und der Blickrichtung verwoben. Manche Quellen betrachten sie daher gemeinsam (Knapp et al. 2013). Aus Gründen der Übersicht unterscheiden wir hier allerdings zwischen dem Bewegungsvorgang der Geste und der eher statischen Körperhaltung, welche an deren Anfang oder Ende steht.

*3.3.1 Funktion der Körperhaltungen*

Eine aufgerichtete Haltung, entweder des Kopfes oder des gesamten Körpers, hängt stark mit der Dominanz und Autorität einer Person zusammen. Beispielsweise konnten Johal et al. (2015) zeigen, dass geschlossene, entspannter wirkende Posen mit gesenktem Kopf weniger Autorität ausstrahlen. Dies galt sowohl für einen humanoiden NAO-Roboter als auch für den Roboter Reeti der Firma

Robopec, welcher lediglich den Kopf bewegen kann.

Die Kopfhaltung spielt auch bei der Darstellung von Emotionen eine wichtige Rolle. Beispielsweise lässt der Roboter Daryl, welcher an der Uni Freiburg entwickelt wurde, zur Darstellung von Trauer oder Enttäuschung den Kopf und die Ohren nach unten hängen (Embgen et al. 2012). Dies soll die Kraftlosigkeit in dieser Gefühlslage darstellen.

Häring et al. stellten bei einer Studie mit dem Roboter NAO fest, dass dessen Körperhaltung die beabsichtigte Emotion deutlicher ausdrücken konnte als dessen Augenfarbe oder von ihm abgespielte Geräusche (Häring et al. 2011).

### 3.3.2 Umsetzung der Körperhaltung

Die Körperhaltungen, welche ein Roboter einnehmen kann, bestehen meistens aus einer einzelnen Schlüsselposition. Ähnlich wie bei Gesten wird definiert, in welcher Position sich die einzelnen Gelenke befinden müssen, damit der Roboter beispielsweise sitzt, stolz aufgerichtet steht oder traurig in sich zusammensackt. Diese Schlüsselposition kann dann als Ausgangslage dienen, aus der heraus eine Geste abgespielt wird und zu der ein Roboter im Anschluss daran zurückkehrt. Beispielsweise gibt es für den Roboter NAO verschiedene vordefinierte Posen, welche er unabhängig von einzelnen Gesten einnehmen kann.[2]

Oft wird die Körperhaltung aber auch direkt mit der Gestik vermischt (Häring et al. 2011; Embgen et al. 2012). Dies hat den Vorteil, dass das Ergebnis leichter vorhersehbar ist.

Zu beachten ist hier, dass Motoren, welche für längere Zeit belastet sind, überhitzen können. Gerade bei Posen wie aufrechtem Stehen oder angehobenen Armen steigt das Risiko dafür. Manche Roboter wie Softbanks Modelle NAO und Pepper besitzen deshalb eine Schutzfunktion, welche beim Erreichen bestimmter Temperaturen veranlasst, dass der Roboter sich hinsetzt oder die Arme hängen lässt, um die Motoren zu entlasten.[3]

## 3.4 Proxemik

Der Begriff „Proxemik" bezeichnet die Lehre von den Abständen, welche Menschen bei der Kommunikation einnehmen (Hall 1963). Jeder Mensch hat bestimmte Vorstellungen davon, welcher Bereich in einer bestimmten Situation ihm oder ihr „gehört", und duldet Eindringlinge in diesem Bereich entsprechend ungern (Knapp et al. 2013). Entsprechend wichtig ist es, dass ein mobiler Roboter sich an die zugehörigen Regeln hält.

### 3.4.1 Funktionen der Proxemik

Die Position und Richtung, welche eine Person im Vergleich zu anderen Menschen einnimmt, kann viel darüber aussagen, wer mit wem zusammengehört und zu welchem Zweck sich diese Personen versammelt haben.

Die Anordnung von Personen, welche sich im Gespräch oder in einer anderen gemeinsamen Aktivität befinden, signalisiert, auf welchen Teil der Umgebung ihre Aufmerksamkeit gerichtet ist (Kendon 2010). Dabei bilden die Körper der Personen üblicherweise eine Grenze zwischen dem Raum, der für die Zusammenarbeit genutzt wird, und dem Bereich, in welchem Außenstehende warten müssen, bis sie in die Gruppe aufgenommen werden. Aus dieser Abgrenzung ergibt sich beispielsweise auch, dass ein Roboter nicht in den Raum zwischen Personen eindringen sollte, die sich gerade im Gespräch miteinander befinden.

Welche Abstände zwischen zwei Personen angemessen sind, hängt von einer Vielzahl von Faktoren ab. Einer davon ist der kulturelle Hintergrund der beteiligten Menschen. So platzieren Menschen aus dem arabischen Kulturkreis beispielsweise zwei Roboter näher aneinander und halten einen kürzeren

---

[2] https://developer.softbankrobotics.com/nao6/nao-documentation/nao-developer-guide/kinematics-data/predefined-postures#naov6-postures
[3] https://developer.softbankrobotics.com/nao6/naoqi-developer-guide/naoqi-apis/naoqi-motion/almotion/diagnosis-effect

Abstand zu diesen ein, als dies bei Deutschen der Fall ist (Eresha et al. 2013).

Neben kulturellen Normen spielt es auch eine wesentliche Rolle, wie vertraut oder sympathisch das Gegenüber ist (Knapp et al. 2013). Beispielsweise konnten Embgen et al. (2012) beobachten, dass der Roboter Daryl beim Ausdruck von Neugierde nicht als aggressiv wahrgenommen wurde, obwohl er dabei die soziale Distanz zu Studienteilnehmern unterschritt. Die Autoren erklären dies mit der langsamen Geschwindigkeit und dem schiefgelegten Kopf, der als freundliches Interesse interpretiert wurde.

### *3.4.2 Umsetzung der Proxemik*

Um sich in geeigneter Weise an einen Menschen anzunähern, muss ein Roboter zuerst bestimmen, wo sich welche Person im Raum befindet. Die meisten Roboter nutzen optische oder ultraschallbasierte Sensoren, um Personen oder Möbelstücke in ihrer Umgebung zu finden und einen sicheren Abstand zu diesen einzuhalten.

Sind die Proxemikvorlieben der betreffenden Person bekannt, können die möglichen Zielpositionen für den Roboter danach sortiert werden, wie ähnlich sie zur bevorzugten Richtung und Distanz sind (Koay et al. 2017). Diese Liste kann der Roboter anschließend nutzen, um die bestmögliche Position zu erreichen, zu welcher der Weg nicht durch Hindernisse versperrt ist.

## 3.5 Mimik

Mimiksignale sind allgegenwärtig. Lächeln oder Stirnrunzeln geben uns Hinweise darauf, was im Kopf einer anderen Person vorgeht, egal ob es ein echter Mensch oder eine Zeichentrickfigur ist. Stilisierte Bilder der verschiedenen Gesichtsausdrücke finden sich in fast jeder Kommunikationsanwendung und auf Social-Media-Plattformen, um das geschriebene Wort in den richtigen emotionalen Zusammenhang zu setzen. Entsprechend naheliegend ist es, auch Roboter mit einem ausdrucksstarken Gesicht auszustatten.

### *3.5.1 Funktionen der Mimik*

Die Verformung der Gesichtszüge wird meistens mit dem Ausdruck von Emotionen in Zusammenhang gesetzt. Als Grundlage dafür dient in vielen Fällen die Arbeit von Ekman und Friesen, welche systematisch die Bewegungen einzelner Gesichtsmuskeln zu prototypischen Emotionsausdrücken zugeordnet haben (Ekman und Friesen 2003, S. Abb. 3). Diese sechs Prototypen entsprechen den Emotionskategorien Überraschung, Furcht, Wut, Ekel, Freude und Trauer, welche oft als „Basisemotionen" bezeichnet werden.

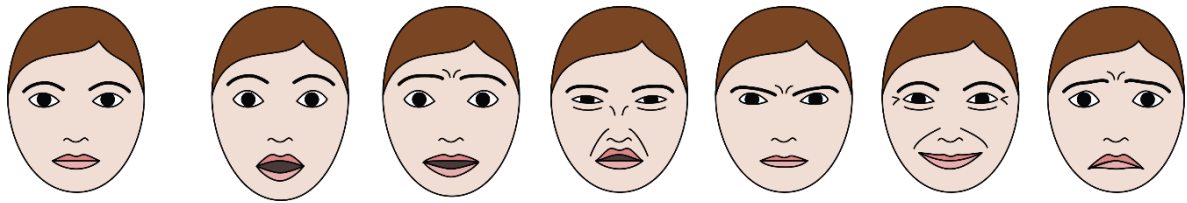

Abb. 3 Prototypische Gesichtsausdrücke für die Basisemotionen

Andere Forscher warnen jedoch davor, dass Studien zu dem Thema durch die verwendeten Methoden verzerrt werden können (Feldman Barrett et al. 2019). Beispielsweise hat es einen erheblichen Einfluss auf die Ergebnisse, ob Probanden echte oder geschauspielerte Gesichtsausdrücke gezeigt werden, und ob sie diesen beliebige Emotionswörter zuordnen dürfen oder aus einer begrenzten Anzahl von Optionen auswählen müssen (Knapp et al. 2013, Feldman Barrett et al. 2019). Zu beachten ist auch, dass die Mimik abhängig von der jeweiligen Person, Situation und Kultur variiert (Feldman Barrett et al. 2019).

Dennoch erleichtert das Einhalten der gleichen Konventionen, auf welche sich Schauspieler und Comiczeichner stützen, menschlichen Beobachtern die Interpretation der Botschaften, welche ein Roboter bezüglich seines inneren Zustands aussendet. Ähnlich wie Ampelfarben in der zugehörigen

Kultur mit positiven oder negativen Situationen assoziiert werden, dienen die Stellung der Mundwinkel oder der Augenbrauen als Symbole dafür, wie der Roboter einen bestimmten Sachverhalt bewertet.

*3.5.2 Umsetzung der Mimik*

Um Gesichtsausdrücke zu beschreiben, wird häufig das „Facial Action Coding System" (FACS) verwendet, welches auf der Aktivierung einzelner Gesichtsmuskeln basiert (Ekman et al. 2002). Im Animationsbereich dient es dazu, Gesichtsausdrücke für Roboter und virtuelle Charaktere nach einem Baukastensystem zusammenzufügen. So lassen sich leicht Varianten oder Mischungen der prototypischen Gesichtsausdrücke darstellen. Ein weiterer Vorteil ist das Vermeiden von Konflikten mit den Sprechbewegungen des Mundes oder dem Blinzeln der Augen, da einzelne Verformungen (sogenannte „Action Units") unabhängig voneinander aktiviert oder deaktiviert werden können.

Für Roboter gibt es grundsätzlich zwei Möglichkeiten, Gesichtsausdrücke zu zeigen. Eine basiert auf der rein grafischen Darstellung des Gesichts, beispielsweise über einen integrierten Bildschirm wie bei dem Roboter Buddy der Firma Bluefrog Robotics. Im Fall des Furhat-Roboters wird ein 3D-animiertes Gesicht auf die Innenseite einer Kunststoffmaske projiziert, um ein möglichst realistisches Ergebnis zu erzielen (Skantze et al. 2014).

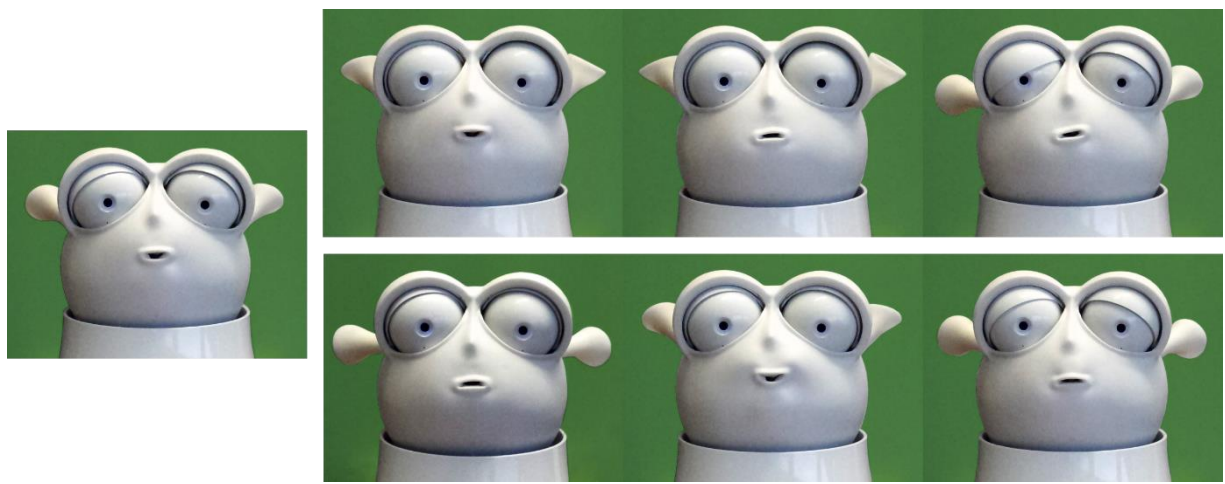

Abb. 4: Prototypische Gesichtsausdrücke dargestellt durch den Roboter Reeti der Firma Robopec.

Die andere Methode ist die physische Verformung von Gesichtsmerkmalen. Dazu werden mechanische Elemente für Augenbrauen, Mundwinkel, Kiefer oder Augenlider von Motoren bewegt. Ein berühmtes Beispiel hierfür ist der Roboter Kismet (Breazeal 2002), welcher am Massachusetts Institute of Technology entwickelt wurde. Für ein weniger mechanisches Aussehen können die Bauteile in einer Plüschhülle oder hinter einer Silikonhaut verborgen werden. Letzteres findet man etwa bei dem Roboter Reeti (siehe Abb. 4) oder bei Sophia von Hanson Robotics.

## 4. Multimodalität

Wie sich in den vorherigen Abschnitten gezeigt hat, sind verbale und nichtverbale Kommunikation oft eng miteinander verflochten. Blickrichtung und Nicken geben dem Sprecher Rückmeldung zu seiner Botschaft, ohne dass der Zuhörer ihm ins Wort fallen muss. Zeige- und bildhafte Gesten übermitteln räumliche Informationen, welche die gesprochene Beschreibung leichter verständlich machen. Ein fröhliches, wütendes oder besorgtes Gesicht zeigt innerhalb von Sekunden, wie der Zuhörer das Gesagte bewertet. Der folgende Abschnitt geht deswegen noch kurz auf das Zusammenspiel zwischen verbaler und nichtverbaler Kommunikation ein.

### 4.1 Zeitliche Koordination von Bewegung und Sprache

Um Bewegungen und Sprache miteinander zu verbinden, gibt es verschiedene Ansätze. In Texte, die von einem Drehbuchschreiber vorgegeben wurden, können beispielsweise Synchronisierungsmarken eingefügt werden (Le und Pelachaud 2012). Sobald der Roboter diese beim Sprechen erreicht, wird auch die zugehörige Bewegung abgespielt. Alternativ lassen sich bestimmte Faustregeln programmieren, etwa dass der Roboter zu Beginn eines Satzes den Kopf zur Seite dreht und am Ende wieder in das Gesicht des Nutzers blickt.

Leider ist es nicht immer einfach, passende Regeln festzulegen. Viele nichtverbale Verhaltensweisen werden von Menschen unterbewusst gezeigt und wahrgenommen und können je nach Situation sehr unterschiedlich ausfallen. Das führt dazu, dass teilweise zu wenig über die zugrunde liegenden Regeln bekannt ist oder eine unübersichtliche Anzahl von Sonderfällen berücksichtigt werden müsste.

Eine mögliche Lösung können hier maschinelle Lernverfahren sein. Dafür müssen zunächst zahlreiche Beispiele aus der Kommunikation zwischen zwei (oder mehr) Menschen gesammelt und mit Anmerkungen versehen werden. Aus diesen Beispielen werden dann statistische Modelle erstellt, welche abhängig von der gegebenen Situation die wahrscheinlichste Aktion auswählen.

### 4.2 Übereinstimmung von Körpersprache und gesprochenem Wort

Für eine konsistente Charakterisierung des Roboters ist es wichtig, dass die verbalen und nichtverbalen Signale übereinstimmen. Wenn die ausgesendeten Botschaften zueinander im Widerspruch stehen, kann dies bei Beobachtern zu Verwirrung und schlimmstenfalls zu Ablehnung führen (Knapp et al. 2013). Widersprüche können als Unehrlichkeit interpretiert werden, entweder weil es so wirkt, als ob die Körpersprache unabsichtlich die „wahren" Gedanken verraten würde, oder weil ein Gesichtsausdruck umgekehrt so aussieht, als wäre er absichtlich aufgesetzt worden (Ekman & Friesen 2003; Knapp et al. 2013).

Unter Umständen lässt sich dies gezielt einsetzen, um beispielsweise Sarkasmus darzustellen. Allerdings muss hierfür berücksichtigt werden, welche Erwartungen die Menschen an den Roboter haben. Auf der einen Seite kann es sein, dass sie dem Roboter derart komplexe Kommunikationsformen nicht zutrauen und den absichtlichen Widerspruch deswegen nicht als solchen erkennen. Andererseits kann es passieren, dass die Menschen mehr in das Verhalten des Roboters hineininterpretieren, als tatsächlich beabsichtigt ist. Daher ist es ratsam, verbales und nichtverbales Verhalten sorgfältig auf einander abzustimmen.

## 5. Zusammenfassung

Bewegung trägt wesentlich dazu bei, einen Roboter „zum Leben zu erwecken", und ist damit eine Grundvoraussetzung für soziale Interaktion. Viele der hier zitierten Studien zeigen, dass die Körpersprache von sozial-interaktiven Robotern ähnlich interpretiert wird wie die von Menschen.

Zeigen oder beschreibende Gesten helfen dabei, Objekte eindeutig zu identifizieren. Ein Blick in die richtige Richtung lässt den Roboter interessiert und aufmerksam wirken, und gezieltes Abwenden des Blicks hilft beim Aushandeln der Sprecherrolle. Die Pose und der Gesichtsausdruck geben Aufschluss darüber, wie der Roboter Ereignisse oder seine Beziehung zum Menschen bewertet. Das Einhalten des richtigen Abstands zeigt schließlich, dass der Roboter die Grenzen des Nutzers respektiert.

Die meisten der hier angesprochenen Verhaltensmuster stellen eigene, umfangreiche Forschungsgebiete dar, welche in diesem Kapitel nur oberflächlich behandelt werden können. Viele Details, wie etwa die konkreten Zusammenhänge zwischen Körpersprache und Emotionen, sind noch nicht restlos geklärt, und Phänomene wie der „Uncanny-Valley"-Effekt werden noch auf längere Sicht Herausforderungen darstellen.

Allerdings trägt die Weiterentwicklung sozial-interaktiver Roboter gleichzeitig dazu bei, das menschliche Verhalten besser zu verstehen. Indem wir Verhaltensmuster auf die Maschinen übertragen, sehen wir umso deutlicher, wo die Gemeinsamkeiten und Unterschiede zur zwischenmenschlichen Kommunikation liegen. Und vielleicht helfen uns die Erkenntnisse daraus nicht nur dabei, uns mit den Maschinen auf Augenhöhe zu verständigen, sondern letztendlich auch mit unseresgleichen.

# 6. Literatur